\pdfoutput=1

\documentclass[11pt]{article}

\usepackage{acl}

\usepackage{times}
\usepackage{latexsym}
\usepackage{graphicx}
\usepackage[T1]{fontenc}

\usepackage[utf8]{inputenc}

\usepackage{microtype}
\usepackage{tabularx}
\usepackage{tablefootnote}
\usepackage{amsmath}
\usepackage{mathrsfs}
\title{SFE-AI at SemEval-2022 Task 11: 
Low-Resource Named Entity Recognition using Large Pre-trained Language Models}

\author{Changyu Hou$^{1}$, Jun Wang$^{1}$\thanks{Corresponding Author.}, Yixuan Qiao$^{1}$, Peng Jiang$^{1}$, Peng Gao$^{1}$, Guotong Xie$^{1*}$ \\
  1 Ping An Healthcare Technology, Beijing, China \\
  \texttt{junwang.deeplearning@gmail.com} \\
  \textbf{Qizhi Lin$^{2}$, Xiaopeng Wang$^{2}$, Xiandi Jiang$^{2}$, Benqi Wang$^{2}$, Qifeng Xiao$^{2}$}\\
  2 Ping An Puhui Enterprise Management Co. Ltd, Shanghai, China \\
  \texttt{wangxiaopeng069@lu.com} \\
  }

\begin{document}
\maketitle
\begin{abstract}

Large scale pre-training models have been widely used in named entity recognition (NER) tasks. However, model ensemble through parameter averaging or voting can not give full play to the differentiation advantages of different models, especially in the open domain. This paper describes our NER system in the SemEval 2022 task11: MultiCoNER. We proposed an effective system to adaptively ensemble pre-trained language models by a Transformer layer. By assigning different weights to each model for different inputs, we adopted the Transformer layer to integrate the advantages of diverse models effectively. Experimental results show that our method achieves superior performances in Farsi and Dutch.  


\end{abstract}

\section{Introduction}
\label{intro}

NER is an essential tool in the application fields of information extraction, question answering system, syntactic analysis, machine translation, etc. It plays a vital role in the process of the practical application of natural language processing technology ~\cite{2018BERT,meng2021gemnet}. 
However, processing complex and ambiguous Named Entities (NEs) is a challenging NLP task in practical and open-domain settings, but has not received sufficient attention from the research community.

Complex NEs, like the titles of creative works (movie/book/song/software names), are not simple nouns and are harder to recognize ~\cite{ashwini2014targetable,fetahu2021gazetteer}. They can take the form of any linguistic constituent, like an imperative clause (“Dial M for Murder”), and do not look like traditional NEs (Person names, locations, organizations). This syntactic ambiguity makes it challenging to recognize them based on their context. Such titles can also be semantically ambiguous, e.g., “On the Beach” can be a preposition or refer to a movie. Finally, such entities usually grow faster than traditional categories, and emerging entities pose yet another challenge. 

In recent years, the deep learning methods have achieved great success in 
NLP (Natural Language Processing). However, some limits need to be addressed. Challenges for deep learning in NLP mainly arise from the scarcity of labeled data.
One way to alleviate the need for large labeled datasets is to pre-train a model on unlabeled data via self-supervised learning, and then transfer the learned model to downstream tasks~\cite{2018BERT}. These methods have been widely applied and have made a massive breakthroughs, such as BERT~\cite{2018BERT}, XLNET~\cite{yang2019xlnet} and GPT~\cite{radford2018improving}.

Pre-trained neural models (e.g., Transformers) have produced high scores on benchmark datasets like CoNLL03/OntoNotes ~\cite{2018BERT}. However, as noted by Augenstein et al. ~\cite{augenstein2017generalisation}, these scores are driven by the use of well-formed news text, the presence of “easy” entities (person names), and memorization due to entity overlap between train/test sets; these models perform significantly worse on complex/unseen entities. Researchers using NER on downstream tasks have noted that a significant proportion of their errors are due to NER systems failing to recognize complex entities ~\cite{luken2018qed,hanselowski2018ukp}.





\begin{figure*}[ht!]
    \centering
    \includegraphics[scale=0.5]{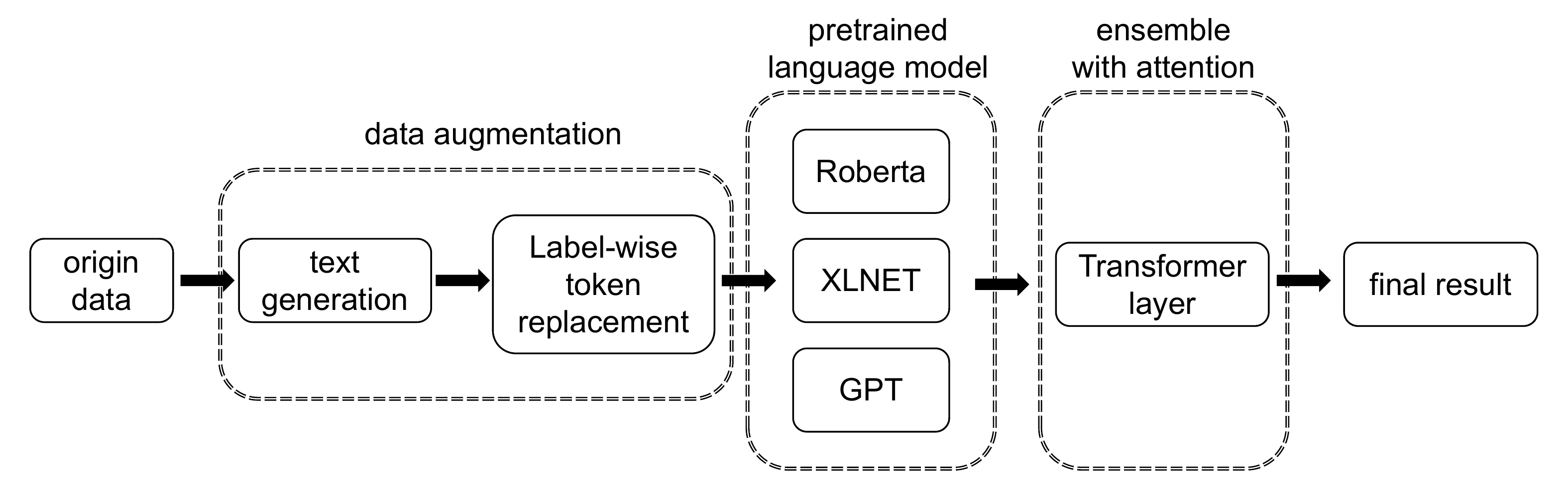}
    \caption{Overview of our architecture for the NER task. The system mainly includes two parts: data augmentation, pre-trained language model ensemble.}
    \label{fig:framework}
\end{figure*}

\section{Related Work}

\textbf{NER.} Named entity recognition (NER) is a sub-task of information extraction, which aims to locate and classify named entities in text into predefined categories. The traditional NER scheme is rule recognition based on manual induction. Based on domain dictionary and grammar rules. Lample~\cite{2016Neural} explored neural structures for NER, in which the bidirectional LSTMs are combined with CRFs with features based on character-based word representations and unsupervised word representations. Despite BiLSTM’s great success in the NER task, it has to compute token representations one
by one, which massively hinders full exploitation of GPU’s parallelism. Therefore, CNN has been proposed to encode words concurrently. In order to enlarge the receptive field of CNNs, iterative dilated CNNs (IDCNN)~\cite{2016Multi} was used.

\noindent\textbf{Pre-training Language Model.} 
Recent large-scale language model pre-training methods such as BERT~\cite{2018BERT}, XLNET~\cite{2021Named} and GPT~\cite{radford2018improving}, further boosted the performance of NER, yielding state-of-the-art performances.

\section{Task Descriptions}

This task challenges NLP enthusiasts to develop complex Named Entity Recognition systems for 11 languages ( In this work, we conducted experiments on Farsi and Dutch). The task focuses on detecting semantically ambiguous and complex entities in short and low-context settings. The task also aims at testing the domain adaption capability of the systems by testing additional test sets on questions and short search queries.

\section{Datasets}



For the entity recognition task~\cite{multiconer-report} on both Farsi and Dutch, there are 15300 pieces of training data. For Farsi, the longest sentence has 46 words, and the shortest sentence has only 1 word, with an average length of 18 words. For Dutch, the longest sentence has 47 words, and the shortest sentence has 2 words, with an average length of 15 words. This means that the length of the current corpus is too short, which means that it is difficult to use a large number of sentence models to cover a large proportion of the length of the current corpus.

In both tasks, there are six types of entities~\cite{multiconer-data}, CW (Creative Work), PER (Person), LOC (Location), GRP (Group), CORP (Corporation), PROD (Product). The distribution of entity types is relatively average. The lowest proportion is PROD, accounting for 12.9\%, and the highest proportion is LOC, accounting for 24.9\%. Many types of entities, short sentence length, lack of sufficient context information, and open-domain information extraction all increase the difficulty of these tasks.


\section{Systems Description}

Overview of our architecture for the NER task was shown in Figure~\ref{fig:framework}. The system mainly includes two parts: data augmentation, pre-trained language model ensemble.

\subsection{Data Augmentation}


For data augmentation, we expanded the existing training data through text generation and label-wise token replacement. Given a sentence s in the training dataset, we aim to generate a new sentence $s^{'''}$ containing the same type entities. Our augmentation consists of the following steps:

1. Entity Mask: Mask all entities in the sentence with specific symbol $[mask]$. We call this modified text $s^{'}$.

2. Sentence Generation: Generate similar sentences using the GPT model. We call this generated sentence $s^{''}$.

3. Entity Infilling: Fill in $[mask]$ tokens with text that has the same entity type. This final output text is called $s^{'''}$.

Besides, we also delete the non-entity part of the sentence randomly to improve the robustness of the model.



\subsection{Model and Benchmarks}
As shown in Figure~\ref{fig:framework}, Roberta, XLNET, and GPT are chosen as the base model in out system. We use these large pre-trained models as the basis and then use the training data to fine-tune each one. After the fine-tune is completed, it is worth mentioning that, the outputs from the last fourth to the last second layer are used to average the parameters as the final output for each model. The reason why we did not use the last layer is to avoid over-fitting.
There is a huge difference between the training dataset and the test dataset (16,000 training data, but 150,000 test data). In order to avoid overfitting on the training set and maintain the generalization ability, we did not use the output of the last layer.

Second, we take the results of all models as the input of the subsequent attention module, as demos in Figure~\ref{fig:framework}.
After obtaining the inputs of base models, we train a transformer layer to assign the corresponding weight to the outputs of different models at the same location, and use the weighted sum of all models as the final output after the model ensemble. Moreover, we add a CRF (Conditional Random Field) layer at the end to limit the final result.

The output of the basic model as shown in below:

\begin{equation}
    \left [
    \begin{array}{ccccc}
    r_1^1     &  ... & r_i^1 & ... & r_L^1\\
    r_1^2     &  ... & r_i^2 & ... & r_L^2\\
    r_1^3     &  ... & r_i^3 & ... & r_L^3\\
    \end{array}
    \right]
\end{equation}
where $r_i$ represents the output of the $i$th model, L means the length of the sentence. Using this matrix as input, we train a transformer layer to learn how to assign different weights to different models. The weight matrix obtained is multiplied by the output results of all models to obtain a matrix with weight, as shown below:

\begin{equation}
    \left [
    \begin{array}{ccccc}
    a_1^1r_1^1     &  ... & a_i^1r_i^1 & ... & a_L^1r_L^1\\
    a_1^2r_1^2     &  ... & a_i^2r_i^2 & ... & a_L^2r_L^2\\
    a_1^3r_1^3     &  ... & a_i^3r_i^3 & ... & a_L^3r_L^3\\
    \end{array}
    \right]
\end{equation}

The vector corresponding to the position can be regarded as the influence of the model on the result of this position, so the final result corresponding to each position can be regarded as the weighted sum of the results of the three models, which is $R_i = [a_i^1r_i^1+a_i^2r_i^2+a_i^3r_i^3]$. This result is used as the final output of the model to predict labels.

\begin{table}[ht!]
\centering
\begin{tabular}{ll}
\hline
Parameter         & value   \\ \hline
sequence length   & 128     \\
batch size        & 24      \\
learning rate     & 0.00003 \\
CRF learning rate & 0.001   \\ 
Dropout           & 0.5     \\
epoch             & 20      \\
\hline
\end{tabular}
\caption{Hyper-parameters of the model.}
\label{tab:parameter}
\end{table}

\subsection{Experimental Setup}

Our implementation is based on PyTorch and we conduct all experiments on one Tesla V100 GPU. Table \ref{tab:parameter} shows the details of the hyper-parameters for our models. The total training epoch is set to 20 and the batch size is set to 24. The initial learning rate is set to $3\times10^{-5}$, and the learning rate of CRF layer is set to $1\times10^{-3}$. Considering the average length of sentences in datesets, the sequence length is 128. Dropout~\cite{2014Dropout} and early-stop are also used in our method to achieve better performance in the open domain.

\begin{table}[]
\centering
\begin{tabular}{lcccc}
\cline{1-4}
                        Model                                              & ACC   & Recall & F1    &  \\ \cline{1-4}
RoBerta                                                               & 0.776 & 0.819  & 0.797 &  \\ 
XLNET                                                                 & 0.791 & 0.816  & 0.804 &  \\ 
GPT                                                                   & 0.799 & 0.764  & 0.781 &  \\ \cline{1-4}
Ensemble model\\ w/o AUG                                                          & 0.816 & 0.822  & 0.819 &  \\ 
Ensemble model\\ w/ AUG
                & \textbf{0.823} & \textbf{0.834}  & \textbf{0.828} &  \\ \cline{1-4}
\end{tabular}
\caption{Benchmarks of Farsi on the dev set. The best performance is highlighted in bold.  w/ AUG and w/o AUG denote model with and without data augmentation.}
\label{tab:farsi}
\end{table}

\begin{table}[ht!]
\centering
\begin{tabular}{lcccc}
\cline{1-4}
                        Model                                              & ACC   & Recall & F1    &  \\ \cline{1-4}
RoBerta                                                               & 0.894 & 0.892  & 0.893 &  \\ 
XLNET                                                                 & 0.892 & 0.899  & 0.895 &  \\ 
GPT                                                                   & 0.882 & 0.880  & 0.881 &  \\ \cline{1-4}
Ensemble model \\w/o AUG                                                          & 0.904 & 0.909  & 0.906 &  \\ 
Ensemble model \\w/ AUG
                & \textbf{0.910} & \textbf{0.911}  & \textbf{0.910} &  \\ \cline{1-4}
\end{tabular}
\caption{Benchmarks of Dutch on the dev set. The best performance is highlighted in bold. w/ AUG and w/o AUG denote model with and without data augmentation.}
\label{tab:dutch}
\end{table}

\section{Results and Discussion}
\label{sec:results}

The results on both Farsi and Dutch data-sets are shown in Table \ref{tab:farsi} and Table \ref{tab:dutch}. Notably, our architecture with data augmentation and model ensemble has substantially improved performance. We also conducted the ablation study of different data augmentation ratios. We found that the best results can be obtained when the data is augmented with a ratio 2 + by three times (as shown in Table \ref{tab:farsi-aug}). Moreover, we also carried out experiments on whether post-processing can improve the performances. The post-processing was conducted as follows: based on 
the dictionary of the training set, we generated entity labels by exact string matching, where conflicted matches were resolved by maximizing the total number of matched tokens. Unfortunately, the final results demonstrated a relatively decreased performance about 0.4 percent. The reason lies in case-sensitive. In many cases, the case of letters plays a vital role in identifying entities.
For example, 'Wat is' is a very prevalent query in Dutch, but 'IS' also represents an organization (Islamic State of Iraq and al-Sham). 

\begin{table}[ht!]
\centering
\begin{tabular}{lcccc}
\cline{1-4}
Augmentation Ratio                                             & ACC   & Recall & F1    &  \\ \cline{1-4}
0 +                                                               & 0.769 & 0.811  & 0.789 &  \\ 
1 +                                                                & 0.772 & 0.811  & 0.791 &  \\ 
2 +                                                          & \textbf{0.776} & \textbf{0.819}  & \textbf{0.797} &  \\ 
3 +
                & 0.775 & 0.816  & 0.795 &  \\ \cline{1-4}
\end{tabular}
\caption{Ablation study of different data augmentation ratios on the dev set  of Farsi using RoBerta.}
\label{tab:farsi-aug}
\end{table}

\section{Conclusions}


This study describes an effective NER system in the SemEval 2022 task 11: MultiCoNER, the experimental results demonstrated the effectiveness of our ensembled large pre-trained language models on low-resource Named Entity Recognition. In future work, we would further improve the performances by exploiting a more complicated ensemble strategy with more diversified models.


\bibliography{anthology,custom}

\begin{thebibliography}{15}
\expandafter\ifx\csname natexlab\endcsname\relax\def\natexlab#1{#1}\fi

\bibitem[{Ashwini and Choi(2014)}]{ashwini2014targetable}
Sandeep Ashwini and Jinho~D Choi. 2014.
\newblock Targetable named entity recognition in social media.
\newblock \emph{arXiv preprint arXiv:1408.0782}.

\bibitem[{Augenstein et~al.(2017)Augenstein, Derczynski, and
  Bontcheva}]{augenstein2017generalisation}
Isabelle Augenstein, Leon Derczynski, and Kalina Bontcheva. 2017.
\newblock Generalisation in named entity recognition: A quantitative analysis.
\newblock \emph{Computer Speech \& Language}, 44:61--83.

\bibitem[{Devlin et~al.(2018)Devlin, Chang, Lee, and Toutanova}]{2018BERT}
J.~Devlin, M.~W. Chang, K.~Lee, and K.~Toutanova. 2018.
\newblock Bert: Pre-training of deep bidirectional transformers for language
  understanding.

\bibitem[{Fetahu et~al.(2021)Fetahu, Fang, Rokhlenko, and
  Malmasi}]{fetahu2021gazetteer}
Besnik Fetahu, Anjie Fang, Oleg Rokhlenko, and Shervin Malmasi. 2021.
\newblock {Gazetteer Enhanced Named Entity Recognition for Code-Mixed Web
  Queries}.
\newblock In \emph{Proceedings of the 44th International ACM SIGIR Conference
  on Research and Development in Information Retrieval}, pages 1677--1681.

\bibitem[{\.F\.Yu and Koltun(2016)}]{2016Multi}
\.F\.Yu and V.~Koltun. 2016.
\newblock Multi-scale context aggregation by dilated convolutions.

\bibitem[{Hanselowski et~al.(2018)Hanselowski, Zhang, Li, Sorokin, Schiller,
  Schulz, and Gurevych}]{hanselowski2018ukp}
Andreas Hanselowski, Hao Zhang, Zile Li, Daniil Sorokin, Benjamin Schiller,
  Claudia Schulz, and Iryna Gurevych. 2018.
\newblock Ukp-athene: Multi-sentence textual entailment for claim verification.
\newblock \emph{arXiv preprint arXiv:1809.01479}.

\bibitem[{Lample et~al.(2016)Lample, Ballesteros, Subramanian, Kawakami, and
  Dyer}]{2016Neural}
G.~Lample, M.~Ballesteros, S.~Subramanian, K.~Kawakami, and C.~Dyer. 2016.
\newblock Neural architectures for named entity recognition.

\bibitem[{Luken et~al.(2018)Luken, Jiang, and de~Marneffe}]{luken2018qed}
Jackson Luken, Nanjiang Jiang, and Marie-Catherine de~Marneffe. 2018.
\newblock Qed: A fact verification system for the fever shared task.
\newblock In \emph{Proceedings of the First Workshop on Fact Extraction and
  VERification (FEVER)}, pages 156--160.

\bibitem[{Malmasi et~al.(2022{\natexlab{a}})Malmasi, Fang, Fetahu, Kar, and
  Rokhlenko}]{multiconer-data}
Shervin Malmasi, Anjie Fang, Besnik Fetahu, Sudipta Kar, and Oleg Rokhlenko.
  2022{\natexlab{a}}.
\newblock Multiconer: a large-scale multilingual dataset for complex named
  entity recognition.

\bibitem[{Malmasi et~al.(2022{\natexlab{b}})Malmasi, Fang, Fetahu, Kar, and
  Rokhlenko}]{multiconer-report}
Shervin Malmasi, Anjie Fang, Besnik Fetahu, Sudipta Kar, and Oleg Rokhlenko.
  2022{\natexlab{b}}.
\newblock Semeval-2022 task 11: Multilingual complex named entity recognition
  (multiconer).
\newblock In \emph{Proceedings of the 16th International Workshop on Semantic
  Evaluation (SemEval-2022)}. Association for Computational Linguistics.

\bibitem[{Meng et~al.(2021)Meng, Fang, Rokhlenko, and Malmasi}]{meng2021gemnet}
Tao Meng, Anjie Fang, Oleg Rokhlenko, and Shervin Malmasi. 2021.
\newblock {GEMNET: Effective gated gazetteer representations for recognizing
  complex entities in low-context input}.
\newblock In \emph{Proceedings of the 2021 Conference of the North American
  Chapter of the Association for Computational Linguistics: Human Language
  Technologies}, pages 1499--1512.

\bibitem[{Radford et~al.(2018)Radford, Narasimhan, Salimans, and
  Sutskever}]{radford2018improving}
Alec Radford, Karthik Narasimhan, Tim Salimans, and Ilya Sutskever. 2018.
\newblock Improving language understanding by generative pre-training.

\bibitem[{Srivastava et~al.(2014)Srivastava, Hinton, Krizhevsky, Sutskever, and
  Salakhutdinov}]{2014Dropout}
Nitish Srivastava, Geoffrey Hinton, Alex Krizhevsky, Ilya Sutskever, and Ruslan
  Salakhutdinov. 2014.
\newblock Dropout: A simple way to prevent neural networks from overfitting.
\newblock \emph{Journal of Machine Learning Research}, 15(1):1929--1958.

\bibitem[{Yan et~al.(2021)Yan, Jiang, and Dang}]{2021Named}
Rongen Yan, Xue Jiang, and Depeng Dang. 2021.
\newblock Named entity recognition by using xlnet-bilstm-crf.
\newblock \emph{Neural Processing Letters}, (1).

\bibitem[{Yang et~al.(2019)Yang, Dai, Yang, Carbonell, Salakhutdinov, and
  Le}]{yang2019xlnet}
Zhilin Yang, Zihang Dai, Yiming Yang, Jaime Carbonell, Russ~R Salakhutdinov,
  and Quoc~V Le. 2019.
\newblock Xlnet: Generalized autoregressive pretraining for language
  understanding.
\newblock \emph{Advances in neural information processing systems}, 32.

\end{thebibliography}
\bibliographystyle{acl_natbib}

\end{document}